\documentclass[conference,a4paper]{IEEEtran}
\IEEEoverridecommandlockouts
\usepackage{cite}
\usepackage{amsmath,amssymb,amsfonts}
\usepackage{algorithmic}
\usepackage{graphicx}
\usepackage{textcomp}
\usepackage{xcolor}

\usepackage{balance}
\usepackage{subfig}
\usepackage{tabularx}
\usepackage{bm}

\def\BibTeX{{\rm B\kern-.05em{\sc i\kern-.025em b}\kern-.08em
    T\kern-.1667em\lower.7ex\hbox{E}\kern-.125emX}}
\begin{document}

\title{Gaussian Mixture Marginal Distributions for Modelling Remaining Pipe Wall Thickness of Critical Water Mains in Non-Destructive Evaluation\\
\thanks{This work is an outcome from the Operationalisation project funded by SydneyWater
Corporation.}
}

\author{\IEEEauthorblockN{Linh Nguyen, Jaime Valls Miro, Lei Shi and Teresa Vidal-Calleja}
\IEEEauthorblockA{\textit{Centre for Autonomous Systems} \\
\textit{University of Technology Sydney}\\
Sydney, Australia \\
\{vanlinh.nguyen, jaime.vallsmiro, lei.shi-1, teresa.vidalcalleja\}@uts.edu.au}
%\and
%\and
%\and
%\and
}

\maketitle

\begin{abstract}
Rapidly estimating the remaining wall thickness (RWT) is paramount for the non-destructive condition assessment evaluation of large critical metallic pipelines. A robotic vehicle with embedded magnetism-based sensors has been developed to traverse the inside of a pipeline and conduct inspections at the location of a break. However its sensing speed is constrained by the magnetic principle of operation, thus slowing down the overall operation in seeking dense RWT mapping. To ameliorate this drawback, this work proposes the partial scanning of the pipe and then employing Gaussian Processes (GPs) to infer RWT at the unseen pipe sections. Since GP prediction assumes to have normally distributed input data - which does correspond with real RWT measurements - Gaussian mixture (GM) models are proven in this work as fitting marginal distributions to effectively capture the probability of any RWT value in the inspected data. The effectiveness of the proposed approach is extensively validated from real-world data collected in collaboration with a water utility from a cast iron water main pipeline in Sydney, Australia. %where the obtained results are highly meaningful for water utility.
\end{abstract}

%Hence, in order to convert the non-Gaussian measurements collected by the sensors to the normally distributed data,


\section{Introduction}
\begin{figure}[t]
\centering
\subfloat[]{\label{3D}
\includegraphics[scale=0.52]{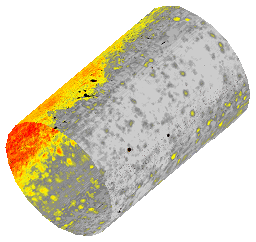}} \hspace*{1.5em}
\subfloat[]{\label{2D}
\includegraphics[scale=0.42]{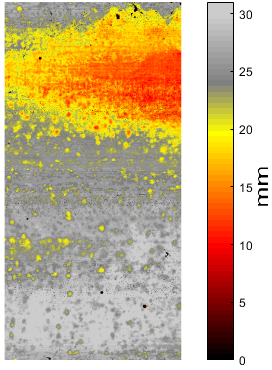}}
\caption{Water pipe remaining wall thickness interpretations in (a) 3D and (b) 2D.}
\label{fig1}
\end{figure}

Approximate 70 percent of the global urban water main assets are buried pipes, and in Australia most of them are now reaching over 100 years old \cite{Miro2014}. Due to deterioration with age, those ageing water pipes are frequently burst, which leads to severe implications for the society and the efficacy and sustainability of water services, including disruption to water supply, obstruction to traffic and damage to properties surrounding the main failures \cite{Vitanage2014}. To prevent the aforesaid severities to community, fully understanding present conditions of the critical water pipes is crucial so that water utilities can have a better pipe-management program. In other words, to effectively forecast remaining lifetime of whole the ageing pipe or even a pipe section, its remaining wall thickness (RWT) needs to be known as can be seen an exemplary example in Fig. \ref{fig1}, where a critical patch is highly visible.

Given a request from Sydney Water Corporation, a water utility in Australia, University of Technology Sydney has developed a non-destructive evaluation/testing (NDE/NDT) system called the rapid response thickness tool (R2T2) that can rapidly assess conditions of a cement lined cast iron (CI) water main pipe. Though the CI pipe wall is invisible since its internal surface is cemented to isolate it from internal corrosion, R2T2 can effectively inspect RWT of the pipe by the use of the NDT techniques \cite{Munoz2017} such as the pulsed Eddy current (PEC) sensing technology \cite{Ulapane2017}. Nevertheless, since the sensing is based on a magnetism principle, where the magnetic field takes time to penetrate through the material (i.e. CI), the tool is constrained by its inspecting speed. More particularly, the primary objective of developing R2T2 is to deploy it into a water pipe at the bursting point when a failure occurs so that it can assess conditions of the pipe in the vicinity to prevent it from subsequent failures in the same area in the near future. Since the water utility aims to minimize long disruption of water supply to customers, the tool is only allowed to be deployed during a short time interval from the bursting to its repair; then the more quickly R2T2 can scan, the longer the pipe section is investigated.

To enhance the issue, it is proposed to scan a part of the pipe and then employ a machine learning  regression technique like Gaussian process (GP) \cite{Rasmussen2006, Nguyen2017a} to predict the rest. Nonetheless, GP is assumed that its input data is normally distributed when using a constant mean, which is not really practical since most of data in nature is non-Gaussian \cite{Nguyen2016b}. To address this issue, it is proposed to exploit a marginal distribution, which marginalize thickness at a location, to model the raw RWT data from R2T2's sensors. That is, the raw RWT readings can be converted to the standard normally distributed data that is represented as input for a GP model. The predictions at the unmeasured locations as outputs of the GP model are then converted to the expected RWT values by the use of the marginal distribution parameters.

\begin{figure*}[t]
\centering
\subfloat[]{\label{r2t2}
\includegraphics[scale=0.35]{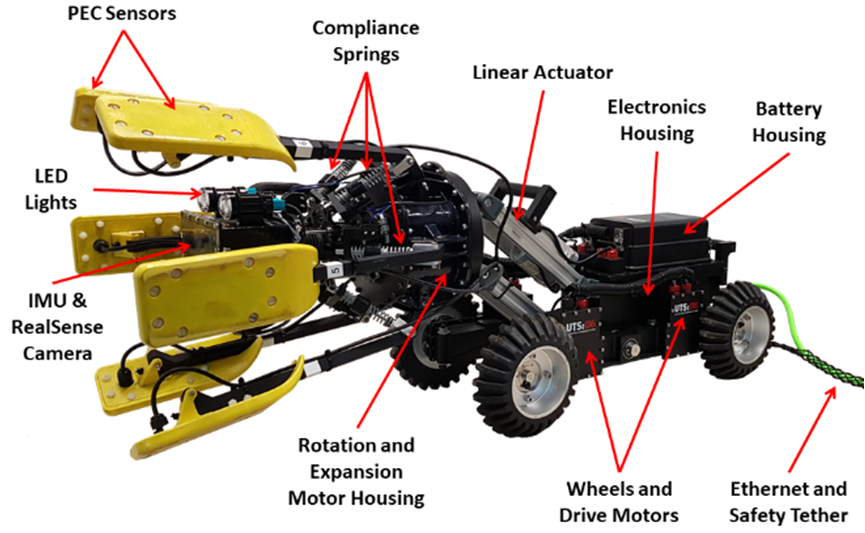}} \hspace*{1.5em}
\subfloat[]{\label{r2t2_deploy}
\includegraphics[scale=0.75]{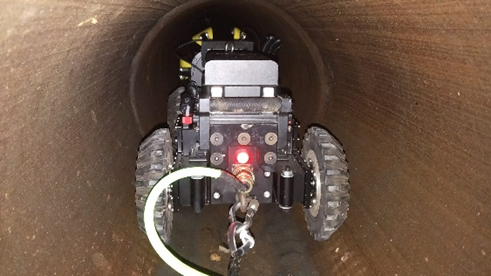}}
\caption{Rapid response thickness tool system \cite{Hunt2018} (a) and its realistic deployment (b).}
\label{fig2}
\end{figure*}

A marginal distribution presents probability of a raw RWT value at an arbitrary point on a pipe surface can be found in literature. For instance, the extreme value (EV) distributions including Gumbel and Weibull are widely used to marginally present the statistics of RWT of a pipe \cite{Asadi2017, Benstock2017}. Due to the tail behaviour, the EV distributions are favoured to model the occurrence of extreme values in corrosion patches such as the minimum RWT or maximum pit depth. Nonetheless, in the cases where the corrosion patches are small but deep, as can be seen in the RWT maps in Figures \ref{p1gt} and \ref{p2gt}, these single-component distributions cannot fit well to the data, which will be discussed further in the following sections. Hence, in this work, we propose to employ Gaussian mixture (GM) as the marginal distribution for modelling the raw PEC sensor readings. Since the GM model is formed by multiple Gaussian components, it is shown to efficiently present probability of having a RWT value at an arbitrary spot on the water pipe. 

The remaining of the paper is organized as follows. Section \ref{sec_2} introduces the inspection tool and how the data was collected in the field before modelling RWT is discussed in Section \ref{sec_3}. Section \ref{sec_4} demonstrates effectiveness of the proposed approach while Section \ref{sec_5} draws conclusions.

\section{Rapid Response Thickness Tool System and Field Data Collection}
\label{sec_2}
To evaluate effectiveness of the Gaussian mixture based marginal distribution on modelling RWT of critical water mains, we implemented the distribution in the data collected by a NDT tool on a realistic water pipe in Sydney, Australia. This section will briefly delineate that tool \cite{Hunt2018} and how the data was collected.
\subsection{Rapid Response Thickness Tool System}

The rapid response thickness tool (R2T2) \cite{Hunt2018} as demonstrated in Fig. \ref{r2t2} is aimed to inspect the CI water main pipes with diameters ranging from 350 mm to 750 mm, which are dominant in the water distribution network in Sydney, Australia. The tool has six arms designed imitating the umbrella configuration, which enables it to adaptively work in the multiple different diameter CI water pipes as aforesaid. On each arm of R2T2, a PEC based elliptical sensor is embedded, which is proved to efficiently sense a CI pipe wall up to 20 mm through a lift-off \cite{Ulapane2017,Ulapane2018}. It is noted that the CI water main pipes in the water distribution network in Sydney have an approximately 10 mm cement lining, which was made to prevent the internal surfaces from corrosion and is considered as a lift-off with respect to the sensor. Nonetheless, in many pipes where the cement lining was applied in-situ, this lift-off can vary from 2 mm to 25 mm, subject to a crown or bottom location. Moreover, given magnetism principle, the sensor cannot measure the wall thickness at a single point but be able to estimate an average thickness under its footprint with a crossing section area, e.g. 50 mm $\times$ 50 mm. The total time for the sensor to comprehensively read one thickness measurement is approximate 150 msec, from emitting a pulsed signal from the excitation coil of the sensor to interpreting a sensor voltage output into the thickness value exploiting our proposed decay curve based algorithms \cite{Nguyen2017b,Ulapane2019}. 

Since the sensor is designed based on the principle of magnetism where the magnetic field takes time to penetrate through the material, assessing the wall thickness of the non-destructive CI water mains is deemed to be quite slow. For instance, R2T2 requires 1.5 hours of scanning to produce a full coverage thickness map of a 20 m section of a 450 mm diameter CI water pipe \cite{Hunt2018}. Unfortunately, the time gap for R2T2 to be implemented into a water pipe at each its failure is very momentary since water utilities would prevent a long disruption from supplying water to their customers. Therefore, if RWT of a CI pipe can be accurately modelled from limited data, R2T2 does not need to scan full coverage of an internal surface of a main but a small part; and the unmeasured part of the main can be efficiently predicted based on the learned model.

\subsection{Field Data Collection}
R2T2 was deployed into an ageing CI water main pipe in the Sydney Water network, Sydney, Australia that was locally burst on 25$^{th}$ May 2019, as shown in Fig. \ref{r2t2_deploy}. The tool was given approximately 2 hours to fully scan the internal surface of a 50 m long section at one side from the breaking point. R2T2 could run forward and backward, and six sensors directly contacted the internal surface (i.e. cement lining). At each run, there were six lines of RWT produced instantly. Since dimension of the sensor is 50 mm $\times$ 50 mm and the pipe diameter is 450 mm, it required the tool to run 6 times to completely scan the whole internal surface of the water main. Some minor measurements overlap between the $1^{st}$ and $6^{th}$ sensors was removed from post-processing the data.

%It is noted that a CI water main pipe in Sydney, Australia was created from multiple spools, which are a standard joint-to-joint pipe segment. And in the 50 m long section R2T2 inspected, there are nine 5.5 m spools. The tool reported that the minimum and maximum RWT in this section are 5.56 mm and 17.74 mm, respectively though the nominal thickness of the pipe is assumed to be 15 mm. It is noticed that due to being cast over hundred years ago, the original wall thickness of the water main was not completely uniform.

%Out of the nine inspected spools, four are in good condition, where their minimum RWT is at least 80 percent of the nominal value, while three have small corrosion patches with the minimum RWT is about 60 percent of the manufactured one (``small" is relative term used to present comparison between area of the corrosion patch and that of the whole internal surface of the pipe). The remaining two are in bad condition since they have the bigger corrosion patches and their minimum RWT is down to approximate 30 percent of the original value. These two bad spools, as illustrated in Figures \ref{p1gt} and \ref{p2gt}, will be utilized for verification of the proposed approach in this work.

Out of the nine inspected spools, the worst two spools, illustrated in Figures \ref{p1gt} and \ref{p2gt}, have been used in this work to verify the validity of the proposed approach.

\section{Remaining Wall Thickness Modelling}
\label{sec_3}
The objective of this study is the assessment of predicting RWT of the CI water mains at unmeasured locations so that R2T2 can speed up the critical condition assessment in future deployments. To this end, information of the unmeasured RWT values can be learned by capturing the spatial correlations among the inspected data. Mathematically, advanced machine learning and spatial statistic tools like Gaussian processes (GP) \cite{Nguyen2017a} can be employed to learn those spatial correlations and then predict the unmeasured thickness given the collected sensor readings. Details arel discussed in this section.

\subsection{Gaussian Mixture based Marginal Distribution}
\label{sec_31}
In our gathered data as illustrated in Figures \ref{p1gt} and \ref{p2gt}, the histogram representations in Fig. \ref{fig4} show that the RWT distribution has very long but tiny tail. Then both the Gumbel and Weibull distributions can capture the probability of the RWT at its high range but not at its low one. It is noted that two data sets called Pipe 1 and Pipe 2 with the histograms plotted in Fig. \ref{fig4} correspond to the RWT maps demonstrated in Figures \ref{p1gt} and \ref{p2gt}, respectively.

In the condition assessment of a critical water main pipe, accurately quantifying RWT at the low range, particularly the minimum RWT, is paramount importance for effectively predicting the likely failure of the pipe in the near future. Since the EV distributions do not fit well to the long but small tail data, in this work it is proposed to employ GM as a marginal distribution to model RWT of the CI water mains. GM distribution can be specified as follows.
\begin{equation}
p(t)=\sum_{i=1}^N\gamma_i\mathcal{N}(t\vert \mu_i,\sigma_i),
\end{equation}
where $N$ is the number of the components and $\gamma_i$ is the component weight with a constraint $\sum_{i=1}^N\gamma_i=1$. $\mu_i$ and $\sigma_i$ are the mean and standard deviation of the $i^{th}$ component, respectively. Advantage of GM distribution as compared with the single-component EV distributions is that it has multiple components of Gaussian distributions, where each component can model a range of the RWT values. The number of Gaussian components can be optimized by the use of Akaike information criterion (AIC). In the following, we will mathematically show how good the GM based marginal distribution is as compared with the Weibull and Gumbel distributions, using two statistical criteria including Kolmogorov–Smirnov (K-S) statistic test and AIC.

\begin{figure*}[t]
\centering
\subfloat[]{\label{p1cdf}
\includegraphics[width=0.8\columnwidth]{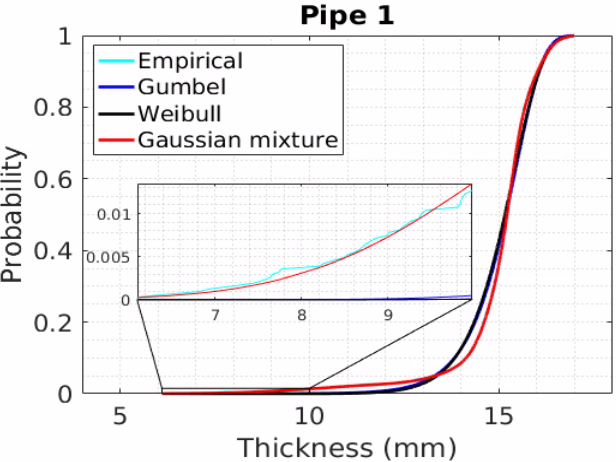}} \hspace*{4em}
\subfloat[]{\label{p2cdf}
\includegraphics[width=0.8\columnwidth]{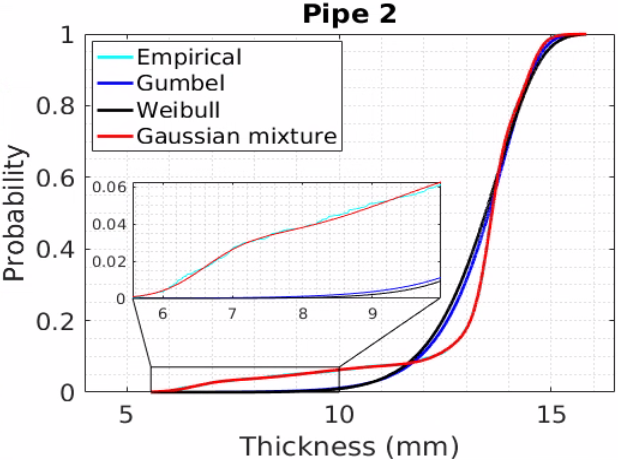}}
\caption{K-S statistic tests on CDF.}
\label{fig3}
\end{figure*}

\begin{figure*}[t]
\centering
\subfloat[]{\label{p1pdf}
\includegraphics[width=0.8\columnwidth]{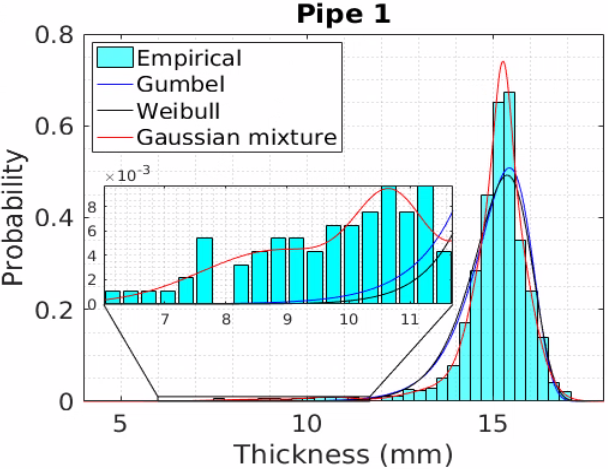}} \hspace*{4em}
\subfloat[]{\label{p2pdf}
\includegraphics[width=0.8\columnwidth]{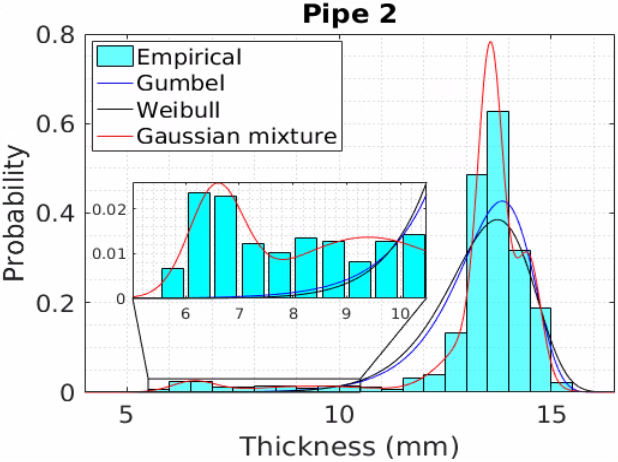}}
\caption{AIC tests on PDF.}
\label{fig4}
\end{figure*}

\subsubsection{K-S statistic test}
Given our collected data as presented in Section \ref{sec_2}, we conducted three different K-S statistic tests with three different null hypothesises, where we hypothesized that one data set (i.e. Pipe 1 or Pipe 2) follows the Gumbel, Weibull and GM distributions, respectively. We first visualize the cumulative distribution functions (CDF) for the both data sets as shown in Fig. \ref{fig3}. It can be seen that at the low range of the RWT, both the Gumbel and Weibull distributions are faraway from the observed data. We then computed the test statistic by
\begin{equation}
KS=\sup_t\vert F_{exp}(t)-F_{obs}(t)\vert,
\end{equation}
where $KS$ is the test statistic while $F_{exp}(t)$ and $F_{obs}(t)$ are the expected hypothesized CDF and the empirical CDF of the RWT values, respectively. Results of the test statistics for both the data sets are summarized in Table \ref{table_1}. Since Pipe 1 has 3080 readings and Pipe 2 has 2968 measurements, at 99\% level of confidence their critical values are approximately given by 0.0294 and 0.0299, respectively. From Table \ref{table_1}, it can be clearly seen that both the null hypothesises of utilizing the Gumbel and Weibull distributions are rejected while the null hypothesis of employing GM is accepted as the marginal distribution for the given data sets.

\begin{table}[thb]
	\renewcommand{\arraystretch}{1.3}
	\caption{K-S STATISTIC TEST RESULTS}
	\label{table_1}
	\centering
	\begin{tabular}{|>{\centering\arraybackslash}m{1cm}|>{\centering\arraybackslash}m{2cm}|>{\centering\arraybackslash}m{2cm}|>{\centering\arraybackslash}m{2cm}|} \hline
		& Gumbel & Weibull & Gaussian mixture  \\ \hline
		Pipe 1 & 0.0715 & 0.0841 & 0.0086  \\ \hline
		Pipe 2 & 0.1448 & 0.1773 & 0.0073  \\ \hline
	\end{tabular}
\end{table}

\subsubsection{AIC}
While the K-S statistic tests focus on the CDF, AIC based comparison is obtained through the probability distribution function (PDF) as can be seen in Fig. \ref{fig4}. Akin to the results obtained in the K-S statistic tests, in those obtained by AIC, GM outperforms both Gumbel and Weibull as the best fit marginal distribution for the given data, especially at the low range of RWT. For numerical comparison, we computed the AIC quantities as follows,
\begin{equation}
AIC=2P-2\log(\mathcal{L}),
\end{equation}
where $P$ is the number of the parameters and $\mathcal{L}$ is the log likelihood of the marginal distribution, respectively. The obtained results for each hypothesized marginal distribution applied for both the data sets are summarized in Table \ref{table_2}. Apparently, GM distribution is best fit to the inspected data as its AIC values are smaller than those of Gumbel and Weibull. It is noticed that it is optimized by 5 and 6 components in the GM marginal distribution for the Pipe 1 and Pipe 2 data sets, respectively.

\begin{table}[thb]
	\renewcommand{\arraystretch}{1.3}
	\caption{AIC RESULTS}
	\label{table_2}
	\centering
	\begin{tabular}{|>{\centering\arraybackslash}m{1cm}|>{\centering\arraybackslash}m{2cm}|>{\centering\arraybackslash}m{2cm}|>{\centering\arraybackslash}m{2cm}|} \hline
		& Gumbel & Weibull & Gaussian mixture  \\ \hline
		Pipe 1 & 7844 & 8053 & 7281  \\ \hline
		Pipe 2 & 9044 & 9677 & 7459  \\ \hline
	\end{tabular}
\end{table}

\subsection{Gaussian Process}
In order to model RWT of a CI water main pipe by employing GP \cite{Nguyen2017a}, since the sensor measures an average thickness value in an area of 50 mm $\times$ 50 mm, it is presumed that the measurement is correspondingly located in the centroid of the sensor footprint.

Let us consider $n$ RWT measurements $\bm{t}=(t_1,t_2,\cdots,t_n)^T\in\mathbb{R}^{n}$ recorded at the locations $\bm{l}=(l_1^T,l_2^T,\cdots,l_n^T)^T\in\mathbb{R}^{n\times 2}$, and those recordings can be statistically modelled by
\begin{equation}
\bm{t}(\bm{l})=\bm{r}(\bm{l})+\varepsilon,
\end{equation}
where $\bm{r}=(r_1,r_2,\cdots,r_n)^T\in\mathbb{R}^{n}$ is the random latent variables at $\bm{l}$. $\varepsilon=(\epsilon_1,\epsilon_2,\cdots\epsilon_n)\in\mathbb{R}^{n}$, where $\epsilon_i$ is an independent identically distributed measurement noise with a zero mean and an unknown variance $\sigma_n^2$. In order to predict RWT at unmeasured positions, it is proposed to model $\bm{r}$ by GP, which maps the probability distribution by the mean and covariance functions. For the purpose of simplicity, it is assumed that the mean function is constant, which averages all the gathered measurements, while the anisotropic automatic relevance determination Matern kernel was selected as the covariance function $C$ in this work. It is noticed that since the pipe is in circumferential cylinder shape, the collected data is periodical. Thus, for the purpose of warping in modelling and predicting of GP, it is proposed to convert a standard two-dimensional location $l_i$ to a periodical four-dimensional location $p_i$ \cite{Miro2018} as follows,
\begin{equation}
p_i=2\pi diag(\lambda^{-1})l_i,
\end{equation}
where $\lambda$ is the vector of the periodical parameters. As a result, the adapted Matern covariance function can be specified by
\begin{equation}
\label{cov_f}
C(p_i,p_j\mid\Theta)=\sigma^2\left(1+\sqrt{3}d\right)\exp\left(-\sqrt{3}d\right),
\end{equation}
where $\sigma$ is the RWT standard deviation and 
\begin{equation}
d=\sqrt{\sum_{k=1}^2\frac{(p_i-p_j)^T(p_i-p_j)}{\eta_k^2}}
\end{equation}
with $\eta_k$ is the characteristic length scale in either the circumferential or longitudinal direction. Note that the hyperparamters $\Theta=(\sigma,\eta_1,\eta,\sigma_n)$ can be obtained by the use of the maximum likelihood technique \cite{Nguyen2016a}.

\subsection{Prediction of Remaining Wall Thickness at Unscanned Pipe Surfaces}
To predict RWT at the $m$ unmeasured locations $\bm{l^*}=((l_1^*)^T,(l_2^*)^T,\cdots,(l_m^*)^T)^T\in\mathbb{R}^{m\times 2}$ (and corresponding periodical positions $\bm{p^*}=((p_1^*)^T,(p_2^*)^T,\cdots,(p_m^*)^T)^T\in\mathbb{R}^{m\times 4}$) given the recorded data $\bm{t}(\bm{p})$, we compute the posterior distribution of the learned GP model with the means $m_{\bm{p^*}}$ and covariances $\Sigma_{\bm{p^*}}$ as follows,
\begin{equation}
m_{\bm{p^*}}\mid\bm{t}(\bm{p})=\mu(\bm{p^*})+\Sigma_{\bm{pp^*}}^T(\Sigma_{\bm{pp}}+\sigma_n^2I)^{-1}\left(\bm{t}(\bm{p})-\mu(\bm{p})\right)
\end{equation}
\begin{equation}
\Sigma_{\bm{p^*}}\mid\bm{t}(\bm{p})=\Sigma_{\bm{p^*p^*}}-\Sigma_{\bm{pp^*}}^T(\Sigma_{\bm{pp}}+\sigma_n^2I)^{-1}\Sigma_{\bm{pp^*}},
\end{equation}
where $\mu(\bm{p^*})$ ($\mu(\bm{p})$) and $\Sigma_{\bm{p^*p^*}}$ ($\Sigma_{\bm{pp}}$) are the vector of means and matrix of covariances at the positions $\bm{p^*}$ ($\bm{p}$), respectively. $\Sigma_{\bm{pp^*}}$ is the matrix of covariances correlating the RWT variables at $\bm{p^*}$ and $\bm{p}$ while $I$ is an identity matrix. All the elements of the covariance matrices are computed by (\ref{cov_f}).

It is noted that the outputs of the GP predictions obey standard normal distribution, to infer the approximately realistic RWT values at the unscanned locations, we conduct the inverse CDF transformation by employing the learned marginal distribution parameters as discussed in Section \ref{sec_31}.

\begin{figure*}[t]
\centering
\subfloat[]{\label{p1dc}
\includegraphics[scale=0.36]{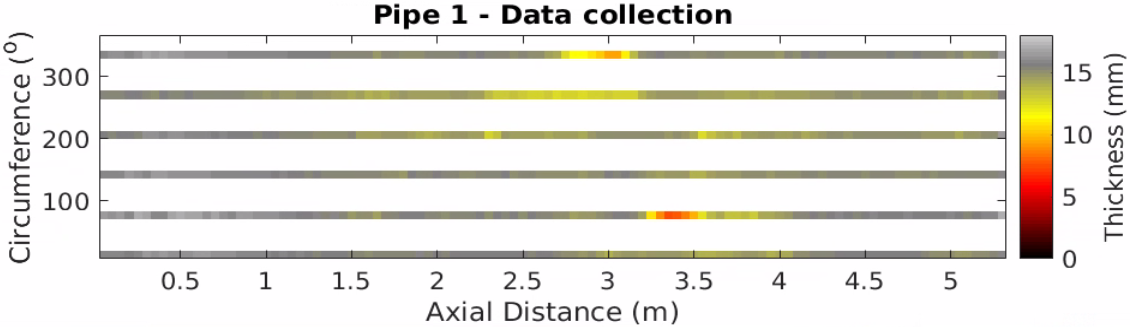}} \hspace*{1.5em} 
\subfloat[]{\label{p2dc}
\includegraphics[scale=0.36]{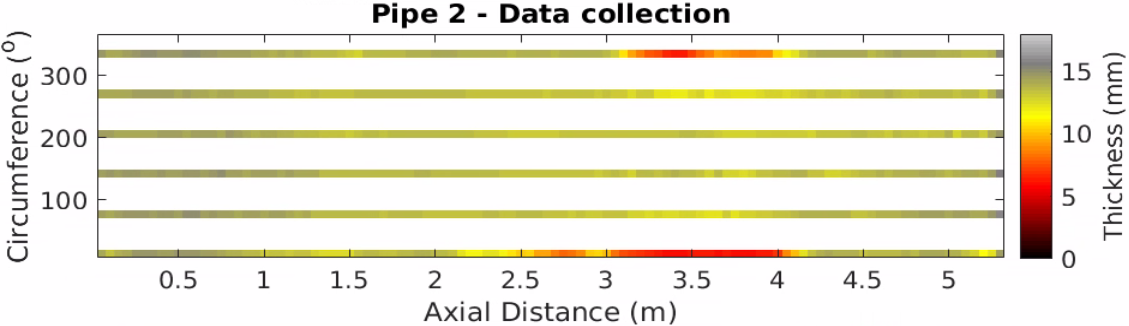}}\\
\subfloat[]{\label{p1pre}
\includegraphics[scale=0.36]{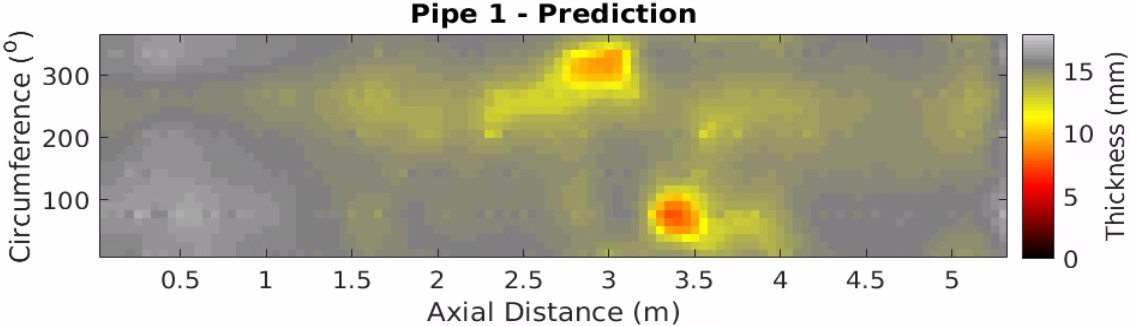}} \hspace*{1.5em}
\subfloat[]{\label{p2pre}
\includegraphics[scale=0.36]{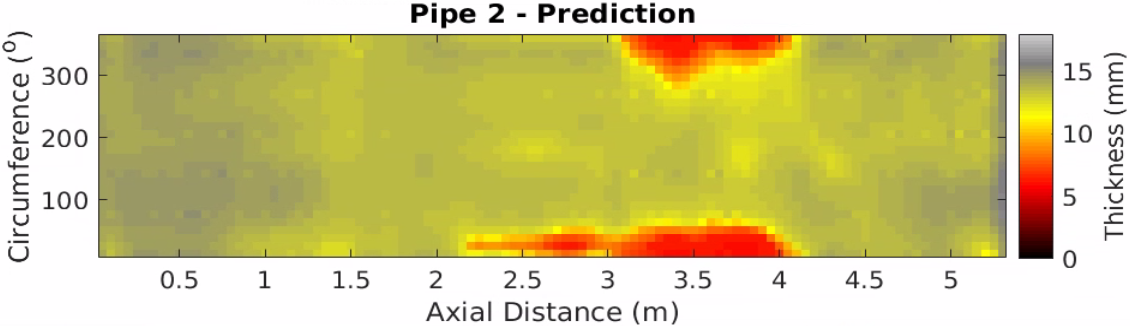}}\\
\subfloat[]{\label{p1gt}
\includegraphics[scale=0.36]{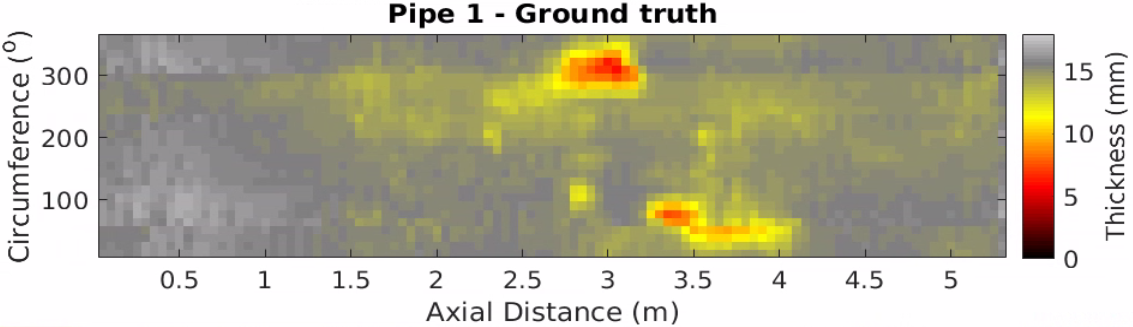}} \hspace*{1.5em}
\subfloat[]{\label{p2gt}
\includegraphics[scale=0.36]{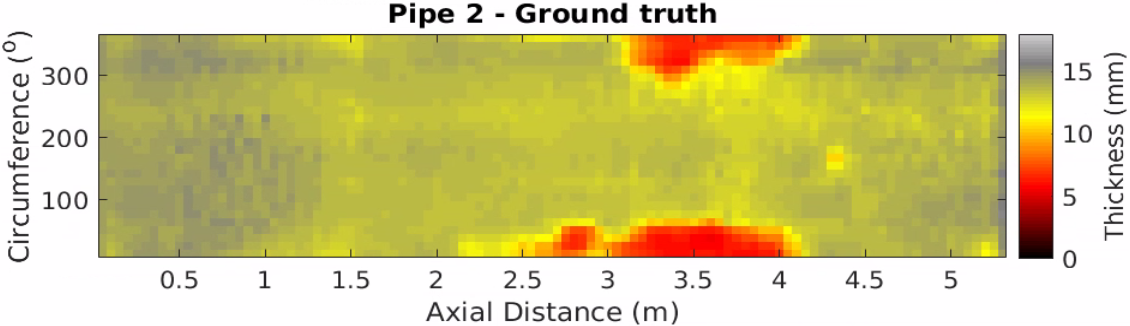}}
\caption{Remaining wall thickness maps: Collected data, predicted results and ground truth. Pipe 1 (left column) and Pipe 2 (right column).}
\label{fig5}
\end{figure*}

\begin{figure*}[t]
\centering
\subfloat[]{\label{p1pdf}
\includegraphics[width=0.76\columnwidth]{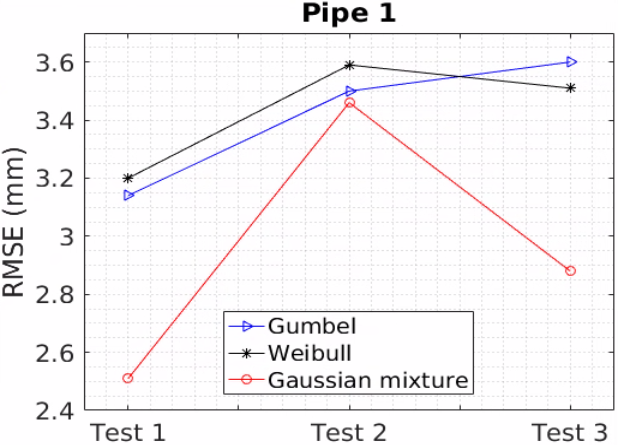}} \hspace*{3em}
\subfloat[]{\label{p2pdf}
\includegraphics[width=0.74\columnwidth]{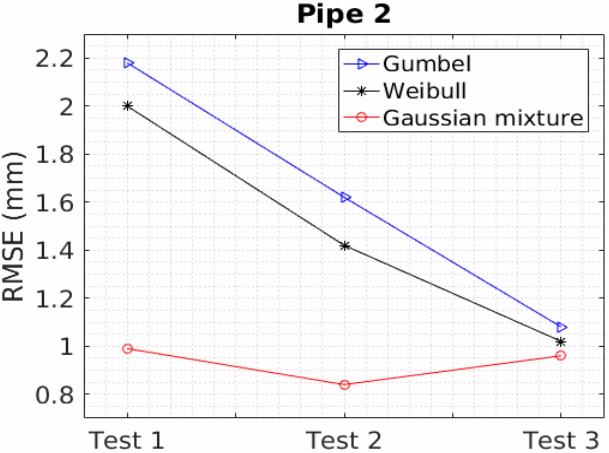}}
\caption{Root mean square errors.}
\label{fig6}
\end{figure*}

\section{Realistic Experimental Results}
\label{sec_4}
To demonstrate effectiveness of the GM based marginal distribution in influencing on the final prediction results, we conducted the experimental tests on the measured data. In other words, the realistic data sets as illustrated in Figures \ref{p1gt} and \ref{p2gt} obtained after R2T2 run 5 times along the water pipe were utilized as the ground truth for comparison. It was presumed that the tool could only run one time along the water main and recorded the data as shown in Figures \ref{p1dc} and \ref{p2dc} for Pipe 1 and Pipe 2, respectively. We expected that the system could predict RWT at the rest of the pipe by filling up the blank areas in Figures \ref{p1dc} and \ref{p2dc}. To this end, a GP model was trained by the sensor readings on each pipe after they were converted from the raw measurements to the standard normal distribution based values by the use of the marginal distribution. The learned GP model was then utilized to predict the corresponding values at the unscanned locations before they were converted to the expected RWT ones based on the marginal distribution parameters. And the predicted RWT results are visualized in Figures \ref{p1pre} and \ref{p2pre} for Pipe 1 and Pipe 2, respectively. It can be clearly seen that the predictions as results of the GP model based on the GM based marginal distribution are highly comparable to the ground truth in Figures \ref{p1gt} and \ref{p2gt}. More importantly, though R2T2 only collected six lines of the data, the prediction could effectively establish the highly accurate corrosion patches as compare to the ground truth, which is crucial in the condition assessment of the critical water main pipes.

For numerical details, we conducted three different tests where R2T2 was supposed to run only one time along the water pipe but the sensor 1 started from the line 1 (the collected data is illustrated in Figures \ref{p1dc} and \ref{p2dc}), the line 2 or line 3, respectively; and the other sensors were shifted accordingly. In each test of each pipe, three GP models were trained by the data converted from the raw measurements by using Gumbel, Weibull and GM as a marginal distribution, respectively. The predicted results as the outputs of the GP models were then converted to the RWT values relied on their corresponding marginal distribution parameters. Differences between the predicted RWT and the ground truth were exploited to compute the root mean square errors (RMSE), which are demonstrated in Fig. \ref{fig6} for the both data sets. The results of the six tests summarized in Fig. \ref{fig6} clearly show outperformance of the GM based marginal distribution as compared with its two counterparts, Gumbel and Weibull. If the RMSE are acceptable, the scanning speed of R2T2 can be ameliorated 5 times faster than the full coverage, which is significantly useful for the condition assessment of the critical CI water main assets.

\section{Conclusions}
\label{sec_5}
The paper has discussed an efficient approach for rapidly inspecting RWT of a metallic water pipeline in NDE. It has been proposed to exploit GM distribution to represent the RWT measurements collected by the PEC sensors from a part of the water main pipe and convert them to the standard normally distributed data for training a GP model. The learned GP model is then utilized to predict the RWT values at the unscanned areas of the main. The results obtained by the proposed method implemented on the real-life dataset show the best fit of the GM distribution to the measurements as compared with the other distributions including Gumbel and Weibull. More importantly, accurately modelling the RWT readings leads to the better RWT predictions, which significantly improves the inspection speed of R2T2.

%\section*{Acknowledgment}
%This publication is an outcome from the Critical Pipes and Sydney Water Operationalisation Projects funded by Sydney Water Corporation, Water Research Foundation of the USA, Melbourne Water, Water Corporation (WA), UK Water Industry Research Ltd, South Australia Water Corporation, South East Water, Hunter Water Corporation, City West Water, Monash University, University of Technology Sydney and University of Newcastle. The research partners are Monash University, University of Technology Sydney and University of Newcastle.

\balance

\bibliographystyle{IEEEtran}
\bibliography{IEEEabrv,References}

\end{document}